\documentclass[runningheads]{llncs}
\usepackage[T1]{fontenc}
\usepackage{graphicx,verbatim}
\usepackage{comment}
\usepackage{float}
\usepackage{amsmath}
\usepackage{multirow}
\usepackage{makecell}
\usepackage{caption}
\usepackage{subcaption}
\usepackage[colorlinks=true,linkcolor=blue,citecolor=blue,urlcolor=blue]{hyperref}
\usepackage{color}

\begin{document}
\title{Graph-Regularized Deep Learning for EEG-Based Emotion Recognition with Psychologically-Grounded Label Structure}
%
\author{Dongyang Kuang \and
Zizheng Ma \and Yushan Zhang \and
Xiaocong Zeng}
\authorrunning{D. Kuang et al.}
%
\institute{School of Mathematics (Zhuhai), Sun Yat-sen University, Guangzhou, China}

\maketitle              
\begin{abstract}

EEG-based emotion recognition is critical for mental health monitoring and affective brain-computer interfaces, yet existing  
deep learning approaches often treat emotion classes as isolated labels, ignoring their psychological interdependencies. We propose a graph-regularized learning framework  
that conceptualizes emotions as nodes in a graph where edges encode proximity based on dimensional emotion theories. We adapt three complementary regularization strategies—Graph  
Label Smoothing (intuitive soft labeling), Commuting distance on graph via Graph Laplacian (spectral graph theory), and Sliced Wasserstein Distance (optimal transport on graph)—ordered 
by increasing computational complexity. These strategies penalize model predictions that deviate from the established emotion topology. 
Our framework is evaluated across three representative backbone architectures: AudioTransformer (pure transformer), Conformer (CNN-transformer hybrid), and DCGNN (causal graph neural network), 
demonstrating architecture-agnostic benefits. Experiments on SEED-IV (4 classes) and SEED-V (5 classes) datasets show consistent improvements: best case up to +5.42\% accuracy and 39\% reduction 
in psychologically implausible misclassifications. Ultimately, our framework help raise the upper bound of performance achievable with standard approaches. Code will be released.


\end{abstract}
\section{Introduction}
Automated emotion recognition from electroencephalography (EEG) signals has substantial clinical applications\cite{schirrmeister2017deep}. In psychiatric care, tracking emotional trajectories over time can inform treatment decisions and 
detect early signs of relapse. Beyond clinical settings, the proliferation of consumer-grade portable EEG devices (e.g., Muse, Emotiv, NeuroSky) has opened opportunities for everyday affective 
computing applications. While deep learning has advanced EEG classification accuracy, standard cross-entropy training treats emotion categories as mutually exclusive and equidistant\cite{Flower2026EEGBasedER}—a 
formulation that contradicts established psychological theory and limits both clinical and consumer applications.

Recent deep learning architectures for EEG emotion recognition, including CNNs \cite{schirrmeister2017deep}, GNNs \cite{song2018eeg}, and transformers \cite{GONG2023104835,GUO2022127700}, focus primarily on feature extraction. They typically employ cross-entropy loss, which treats all misclassifications equally and ignores the psychological structure of emotions. While label smoothing \cite{7780677,bdcc9110285,Maratos2023LabelSF} and hierarchical classification \cite{Silla2011A,11036586,9156542} have been explored, systematically integrating emotion topology into training objectives remains largely unaddressed. This gap is clinically significant: dimensional models \cite{russell1980circumplex} show emotions occupy a structured space where some states are inherently more similar. For instance, confusing Fear with Sadness (both negative valence) is psychologically more plausible than confusing Fear with Happy. Yet, standard losses penalize both errors equally.\par
To address this, we propose an \textbf{emotion graph} modeling the discrete emotion space, where edges encode psychological proximity. This structure defines \textbf{emotion transition difficulty}—the cost of misclassifying emotions—using three principled distance metrics: (1) \textbf{adjacency-based proximity} via Graph Label Smoothing for local relationships; (2) \textbf{commuting distances} via the Graph Laplacian pseudo-inverse for global connectivity; and (3) \textbf{optimal transport distances} via Sliced Wasserstein to measure distributional transformation effort. These metrics quantify prediction error distances, teaching the model that confusing adjacent emotions is more acceptable than distant ones. We also use uncertainty-based adaptive weighting \cite{kendall2018multi} to balance classification and regularization.

Our contributions are: 
1. \textbf{Novel graph-based regularization} for EEG emotion recognition that incorporates psychological proximity into training, helping raise the upper bound of performance achievable with standard approaches and enhance clinical relevance. 
2. \textbf{Architecture-agnostic validation} across three representative backbone families:transformer (AudioTransformer), CNN-transformer hybrid(Conformer), and GNN (DCGNN) backbones on two benchmarks, showing improved best accuracy achievable and better feature representation space. 
3. \textbf{Proximity violation metric} to measure meaningful errors that respect psychological proximity. 

\section{Method}
\subsection{Emotion Graph Construction}
\begin{figure}[tbp]
\centering
\includegraphics[width=1.0\linewidth, height=0.3\linewidth]{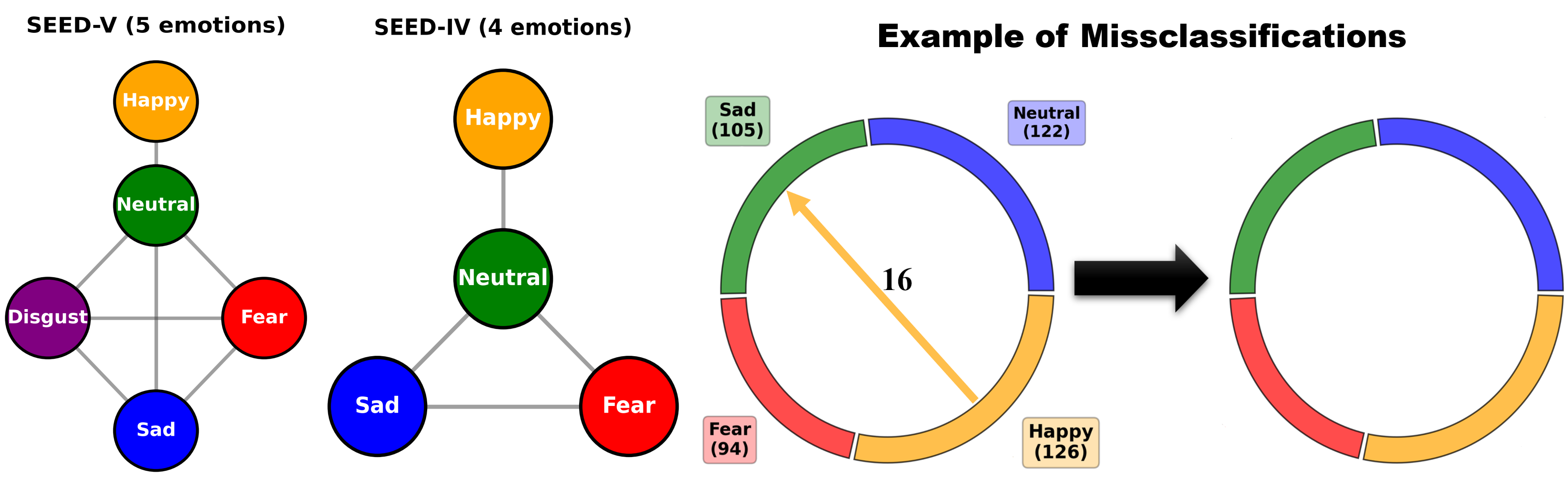}
\caption{Emotion graphs for SEED-V (5 emotions) and SEED-IV (4 emotions) and an example of decreasing more problematic misclassifications after our graph-based regularization, for exmaple, Happy->Sad cases.}
\label{fig:emotion_graphs}
\end{figure}

We construct emotion graphs based on Russell's circumplex model~\cite{russell1980circumplex}, as illustrated in Fig. \ref{fig:emotion_graphs},positioning each emotion in the two-dimensional valence--arousal space and connecting pairs whose coordinates are proximal. Edges therefore encode \emph{psychological proximity}: emotions that are nearby in affective space are adjacent in the graph, so the model is penalized less for confusing them than for confusing affectively distant states.

\textbf{SEED-IV}~\cite{zheng2018emotionmeter} is a widely-used EEG benchmark recorded from 15 subjects across 3 sessions, each containing 72 film-clip trials eliciting 4 emotions: \textit{Neutral}, \textit{Sad}, \textit{Fear}, and \textit{Happy}. 
Signals are captured via 62-channel EEG at 1000\,Hz. The graph topology shown in \ref{fig:emotion_graphs} is justified by valence-arousal coordinates: \textit{Happy} is the \emph{sole positive-valence} emotion in this label set; it shares no direct psychological proximity with the negative-valence states and therefore connects only to \textit{Neutral}, which occupies the valence origin. \textit{Sad} and \textit{Fear} are both negative-valence with partially overlapping arousal levels (low-to-moderate and moderate-to-high, respectively), justifying a direct edge. \textit{Neutral} acts as a topological hub adjacent to all nodes, reflecting its central position in the circumplex. 

\textbf{SEED-V}~\cite{liu2016emotion} extends the paradigm to 5 emotions (\textit{Disgust}, \textit{Fear}, \textit{Sad}, \textit{Neutral}, \textit{Happy}) with 16 subjects and 3 sessions of 15 trials per emotion (75 trials per session), 
also recorded at 62 channels. The richer label set amplifies the importance of emotion topology: the three negative emotions (Disgust, Fear, Sad) form an interconnected cluster reflecting shared negative valence and 
high arousal overlap; Neutral connects to all nodes as an affective origin; \textit{Happy} again remains the \emph{sole positive-valence} emotion, and its exclusive connection to \textit{Neutral} reflects the psychologically motivated absence of any direct path to the negative-valence cluster.

\subsection{Graph-Based Regularization}
Given adjacency matrix $A$, we compute the graph Laplacian $L = D - A$ (where $D$ is the degree pmatrix) and graph distance matrix $C$ via shortest paths. We adapt three complementary 
regularization strategies from different mathematical traditions, presented in order of increasing computational complexity:

\textbf{Graph Label Smoothing (Soft Labeling).} We extend standard label smoothing by distributing probability mass according to graph adjacency rather than uniformly:
\begin{equation}
y_i^{smooth} = (1-\alpha) \cdot y_i + \alpha \cdot \tilde{A}[y_i]
\end{equation}
where $\tilde{A}$ is the normalized adjacency matrix and $y_i$ is the true class. This intuitive approach softens labels toward psychologically adjacent emotions with minimal computational 
overhead.

\textbf{Graph Laplacian Loss with Commuting Distance (Spectral Graph Theory).} Using the pseudo-inverse $H = L^+$, which encodes commuting distances between nodes \cite{fouss2007random}:
\begin{equation}
\mathcal{L}_{GL} = \frac{1}{B}\sum_{i=1}^{B} (\hat{y}_i - y_i)^T H (\hat{y}_i - y_i)
\end{equation}
where $\hat{y}_i$ is the predicted distribution and $y_i$ is the one-hot ground truth. The commuting distance measures the expected number of steps for a random walk to travel between 
two nodes and return, capturing global graph connectivity.

\textbf{Sliced Wasserstein Loss (Optimal Transport).} We compute the Sliced Wasserstein Distance between distributions \cite{bonneel2015sliced} \cite{titouan2019optimal}:
\begin{equation}
\mathcal{L}_{SW} = \frac{1}{L}\sum_{l=1}^{L} W_1(\theta_l^\sharp \hat{y}, \theta_l^\sharp y)
\end{equation}
where $L$ is the number of random projections, $\theta_l$ is a random unit direction sampled uniformly on the unit sphere, $\theta_l^\sharp \hat{y}$ and $\theta_l^\sharp y$ denote the 
one-dimensional pushforward (projection) of the predicted distribution $\hat{y}$ and ground-truth distribution $y$ onto $\theta_l$, and $W_1$ is the one-dimensional Wasserstein-1 distance. 
The graph shortest-path distance serves as the ground metric, so that projections respect the emotion topology. Random slicing reduces the computational cost of full optimal transport 
while preserving sensitivity to the underlying graph structure.

\textbf{Computational Complexity.} Let $K$ denote the number of emotion classes and $B$ the batch size. GLS requires $O(K)$ per sample for adjacency lookup, yielding $O(BK)$ total. GL involves 
matrix--vector multiplication with the $K \times K$ pseudo-inverse, costing $O(BK^2)$. SW requires $L$ random projections and 1D sorting, giving $O(BLK\log K)$ where $L$ is typically 50--100. 
For small $K$ (4--5 classes here), all three methods are computationally efficient, but GLS scales best for larger emotion taxonomies.

Through developed from different perspectives, All the above three configurations repect the emotion dynamics and topology, penalizing more heavily for psychologically implausible confusions (e.g. transition as Happy$\rightarrow$Sad is more difficult than Happy$\rightarrow$Neural). To combine the classification and regularization objectives, we consider two weighting strategies.
With \textbf{fixed weighting}, the regularization strength is controlled by a scalar $\lambda$:
\begin{equation}
\mathcal{L} = \mathcal{L}_{CE} + \lambda \cdot \mathcal{L}_{reg}, \quad \mathcal{L}_{reg} \in \{\mathcal{L}_{GL}, \mathcal{L}_{SW}, \mathcal{L}_{GLS}\}
\end{equation}
where $\lambda \in \{0.05, 0.1, 0.2\}$ is selected by cross-validation. Alternatively, \textbf{adaptive weighting}~\cite{kendall2018multi} learns the balance automatically 
via log-variance parameters $s = \log\sigma^2$, yielding the homoscedastic uncertainty objective:
\begin{equation}\label{eqn:adaptive}
\mathcal{L} = e^{-s_{CE}}\mathcal{L}_{CE} + e^{-s_{reg}}\mathcal{L}_{reg} + s_{CE} + s_{reg}
\end{equation}
This removes the need for manual $\lambda$ selection while providing a principled probabilistic interpretation: the model down-weights tasks it is more uncertain about during training.

\noindent
\textbf{Proximity Violation Metric}
We define the Proximity Violation ratio:
\begin{equation}
PV = \frac{\text{\# \{misclassifications to non-adjacent classes\} }}{\text{\# \{total misclassifications\} }} \times 100\%
\end{equation}

Lower PV indicates predictions that, even when wrong, remain within psychologically plausible emotion neighborhoods—a clinically meaningful property. For example, Happy$\rightarrow$Neutral 
is not a violation (adjacent), whereas Happy$\rightarrow$Sad is (non-adjacent, crossing the valence boundary).

\section{Experiments}
\subsection{Setup}
We evaluate our framework on SEED-IV~\cite{zheng2018emotionmeter} and SEED-V~\cite{liu2016emotion}. The pre-processed differential entropy features across 5 frequency bands are adopted as input features. To demonstrate that our approach is architecture-agnostic, we test it on three backbone models spanning 
distinct deep learning paradigms: AudioTransformer~\cite{gong21b_interspeech}, a pure transformer that captures long-range temporal 
dependencies; Conformer~\cite{Gulati2020ConformerCT}, a CNN-transformer hybrid that combines local feature extraction with global 
attention; and DCGNN~\cite{11048752}, a graph neural network that explicitly models spatial relationships between EEG channels. Rather than competing with existing methods on leaderboard accuracy, this study is specifically designed to evaluate whether graph-based regularization yields consistent, architecture-agnostic improvements. 

For each backbone, we compare a cross-entropy baseline ($\lambda{=}0$) against three regularization variants—Graph Laplacian (GL), Sliced Wasserstein (SW), and Graph Label Smoothing (GLS)—with
fixed weights $\lambda \in \{0.05, 0.1, 0.2\}$ and an adaptive weighting scheme (auto) as introduced in \cite{kendall2018multi}. 
All experiments follow a 10-fold subject-dependent cross-trial protocol (non-overlapping trials per fold), with Accuracy, Macro F1, and PV averaged over all folds and subjects. 
We report the best-epoch performance over 200 training epochs per fold, applied uniformly to both the baseline and all regularization variants. 
Bold denotes the best regularization variant per model--dataset pair.

\subsection{Results}

\begin{table}[t]
\centering
\caption{Test Performance on SEED-IV (4-class) and SEED-V (5-class) datasets. For each backbone and regularization method, we swept the regularization strength over {auto, 0.05, 0.1, 0.2} and report only the best-performing setting in terms of Macro F1 representing the upper bound of achievable performance. Bold denotes the best regularization variant per model--dataset pair.}
\label{tab:combined_lr_split_lambda}
\resizebox{\linewidth}{!}{
\begin{tabular}{|l|l|c|c|c|c|c|c|c|c|}
\hline
\multirow{2}{*}{Model} & \multirow{2}{*}{Reg.} & \multicolumn{4}{c|}{SEED-IV (4-class)} & \multicolumn{4}{c|}{SEED-V (5-class)} \\
\cline{3-10}
 & & $\lambda$ & Acc (\%)$\uparrow$ & F1 (\%)$\uparrow$ & PV (\%)$\downarrow$& $\lambda$ & Acc (\%)$\uparrow$& F1 (\%) $\uparrow$& PV (\%)$\downarrow$\\
\hline
\multirow{4}{*}{\makecell[l]{Audio-\\Transformer}} & None   & -    & 83.12$\pm$7.27          & 81.21$\pm$8.00          & 21.8$\pm$12.1 & --   & 83.06$\pm$6.98          & 74.69$\pm$10.64          & 22.4$\pm$14.2 \\
            & GL & auto & 84.01$\pm$6.83          & 82.31$\pm$7.38          & \textbf{20.6$\pm$10.7} & auto & 83.93$\pm$7.30          & \textbf{76.36$\pm$9.94}  & 21.5$\pm$15.8 \\
            & SW      & auto & \textbf{84.56$\pm$6.69} & \textbf{82.94$\pm$7.03} & 21.5$\pm$13.8 & auto & \textbf{84.24$\pm$7.51} & 76.22$\pm$11.10          & 22.0$\pm$16.6 \\
            & GLS     & auto & 83.94$\pm$6.82          & 82.05$\pm$7.34          & 23.0$\pm$13.8 & auto & 83.26$\pm$7.26          & 75.17$\pm$10.41          & \textbf{20.0$\pm$16.1} \\
\hline
\multirow{4}{*}{Conformer}   & None   & -    & 83.86$\pm$6.49          & 81.73$\pm$7.26          & \textbf{24.0$\pm$9.6} & --   & 84.31$\pm$8.83          & 77.06$\pm$11.60          & 19.6$\pm$16.6 \\
            & GL & auto & \textbf{84.34$\pm$7.18} & \textbf{82.26$\pm$8.11} & 25.6$\pm$11.2 & 0.1  & 89.09$\pm$7.44          & 82.82$\pm$10.99          & 16.1$\pm$14.7 \\
            & SW      & 0.1  & 84.22$\pm$6.62          & 82.28$\pm$7.62          & 27.8$\pm$16.2 & 0.1  & \textbf{89.73$\pm$7.09} & \textbf{82.96$\pm$10.75} & \textbf{11.9$\pm$12.3} \\
            & GLS     & 0.1  & 83.73$\pm$6.81          & 82.00$\pm$7.49          & 26.7$\pm$10.1 & 0.05 & 84.92$\pm$7.89          & 78.04$\pm$10.92          & 21.6$\pm$16.8 \\
\hline
\multirow{4}{*}{DCGNN}       & None   & -    & 72.59$\pm$8.46          & 69.83$\pm$9.05          & 26.3$\pm$12.1 & --   & 74.15$\pm$14.96         & 66.73$\pm$17.05          & 16.9$\pm$14.8 \\
            & GL & auto & \textbf{73.48$\pm$8.06} & \textbf{70.67$\pm$8.90} & 27.1$\pm$11.2 & auto & 75.14$\pm$14.44         & 67.38$\pm$16.39          & 18.1$\pm$15.0 \\
            & SW      & 0.2  & 72.84$\pm$8.30          & 70.01$\pm$8.80          & 26.4$\pm$11.7 & auto & \textbf{75.44$\pm$14.58}& \textbf{67.85$\pm$16.71} & 17.8$\pm$13.9 \\
            & GLS     & 0.05 & 73.07$\pm$7.85          & 70.19$\pm$8.69          & \textbf{24.2$\pm$11.3} & 0.05 & 74.91$\pm$14.33         & 66.97$\pm$16.67          & \textbf{16.6$\pm$14.0} \\
\hline
\end{tabular}
}
\end{table}

As shown in Tables \ref{tab:combined_lr_split_lambda}, incorporating graph-based regularization consistently enhances classification performance across all three backbone architectures. Notably, on SEED-V, Conformer 
with Sliced Wasserstein (SW) regularization achieves up to a \textbf{+5.42\%} increase in accuracy and a \textbf{39\%} reduction 
in Proximity Violations (PV). On SEED-IV, the graph-based DCGNN architecture benefits most dramatically, with accuracy improvements reaching +5\%. Furthermore, the adaptive weighting mechanism (\textbf{auto}) performs 
competitively without requiring manual hyperparameter tuning, proving particularly effective for graph-based models. Crucially, the general reduction in PV indicates that the models learn to make psychologically plausible 
predictions, avoiding severe misclassifications across distant emotion nodes.


\begin{figure}[htbp]
    \centering
    \begin{subfigure}[t]{0.496\textwidth}
        \centering
        \includegraphics[width=\textwidth, height=0.42\linewidth]{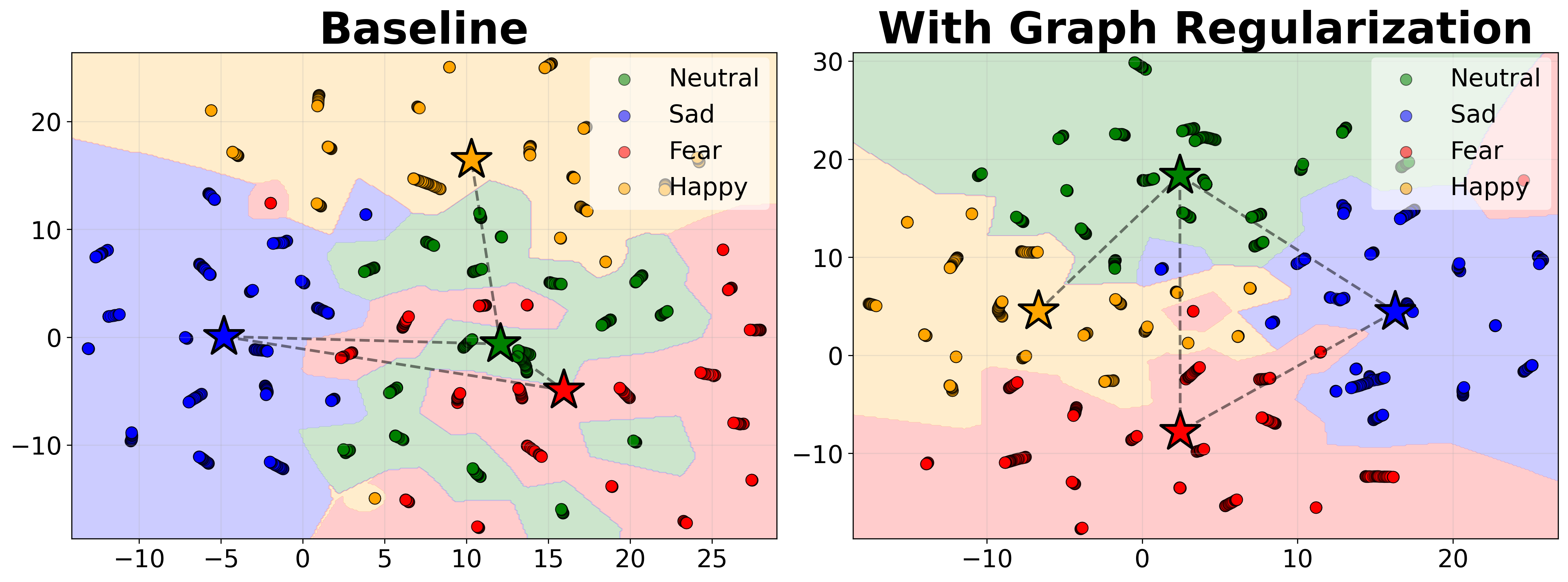}
        \caption{AudioTransformer on SEED-IV}
        \label{fig:sub_a}
    \end{subfigure}
    \hfill
    \begin{subfigure}[t]{0.496\textwidth}
        \centering
        \includegraphics[width=\textwidth, height=0.42\linewidth]{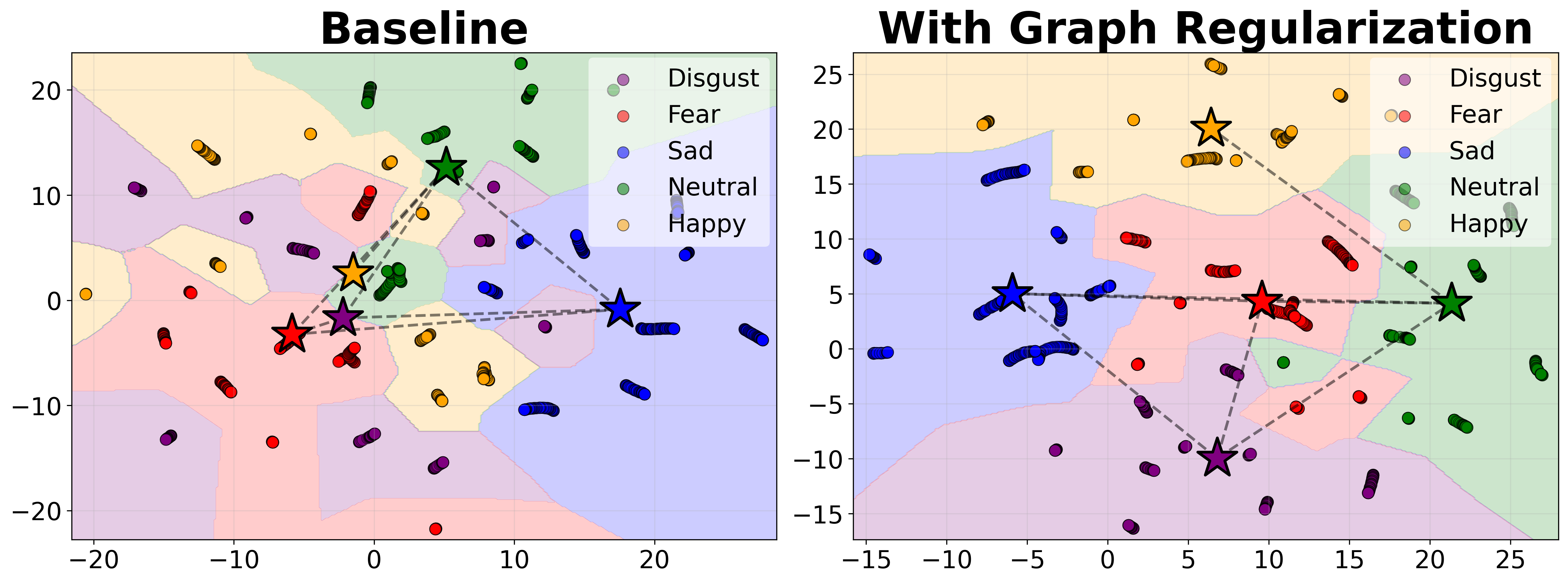}
        \caption{Conformer on SEED-V}
        \label{fig:sub_b}
    \end{subfigure}
    
    \caption{UMAP visualization of learned representations}
    \label{fig:total_figure}
\end{figure}

For better interpretability of the impact of graph-based regularization, we use UMAP to visualize the learned representations as shown in Fig. \ref{fig:total_figure}. 
With SW regularization, the clusters corresponding to different emotion classes become more distinct and well-separated, while cases without regularization show more scattered and overlapping clusters. This indicates that the model learns a more structured representation space that respects the psychological proximity of emotions, leading to fewer implausible misclassifications.
This visualization supports our quantitative findings, demonstrating that graph-based regularization helps the model learn representations that align better with the underlying psychological structure of emotions.

\begin{table}[tbhp]
\centering
\caption{Subject-Level F1 Improvement Rate (\% Subj\,$\uparrow$). \textbf{Bold} values are aggregated across different $\lambda$ values,methods and models. We test them under one-sided Wilcoxon rank-sum test on the hypotheis that whether or not >50\% subjects will show improvements with regularizations without specifying actual $\lambda$ value. (***) denotes $p<0.01$, which confirm that the improvement brought by proposed graph-based regularization is statistically significant.}
\label{tab:seediv_scheme3}\label{tab:seedv_scheme3}
\resizebox{\linewidth}{!}{
\begin{tabular}{|l|l|c|c|c|c|}
\hline
Dataset & Method & AudioTransformer & Conformer & DCGNN & Avg(Method) \\
\hline
\multirow{4}{*}{SEED-IV} & GL & 73.33 & 56.67 & 55.00 & $61.67 \pm 8.28$ \\
 & SW & 76.67 & 55.00 & 51.67 & $61.11 \pm 11.08$ \\
 & GLS & 58.33 & 58.33 & 46.67 & $54.44 \pm 5.50$ \\
\cline{2-6}
 & Avg(Model) & $69.44 \pm 7.97$ & $56.67 \pm 1.36$ & $51.11 \pm 3.42$ & \textbf{59.07 $\pm$ 9.20}(***) \\
\hline
\multirow{4}{*}{SEED-V} & GL & 57.81 & 67.19 & 48.44 & $57.81 \pm 7.65$ \\
 & SW & 62.50 & 73.44 & 57.81 & $64.58 \pm 6.55$ \\
 & GLS & 53.12 & 54.69 & 56.25 & $54.69 \pm 1.28$ \\
\cline{2-6}
 & Avg(Model) & $57.81 \pm 3.83$ & $65.10 \pm 7.80$ & $54.17 \pm 4.10$ & \textbf{59.03 $\pm$ 7.17}(***) \\
\hline
\end{tabular}
}
\end{table}

Table~\ref{tab:seediv_scheme3} reports the percentage of subjects whose individual F1 score improves under regularization (\% Subj\,$\uparrow$) across all method--backbone combinations. For example, a value of 50 indicates that 50\% of subjects experienced an improvement in F1 score compared to the baseline when a specific regularization method was applied. For the entry corresponding to each backbone model and regularization method, the value is already averaged across all tested regularization strengths ($\lambda$ values) for that method, so it reflects the overall likelihood of improvement regardless of the specific regularization strength.                              
This reveals that the benefit of encoding emotion topology is both \emph{quantitatively consistent}---improvements appear across the majority of individual subjects---and 
\emph{qualitatively architecture-agnostic}---the pattern holds regardless of whether the backbone is a pure transformer, a CNN--transformer hybrid, or a graph neural network.

\subsection{Generalization Analysis}

On SEED-IV, all three regularization methods improve over baseline: on average across backbones, GL benefits 61.67\%, SW 61.11\%, and GLS 54.44\% of subjects (overall average 59.07\%, 
Table~\ref{tab:seediv_scheme3}). AudioTransformer shows the largest individual gains, with SW improving 76.67\% of subjects. On SEED-V, improvements are even more pronounced: 
the overall subject-level F1 improvement rate reaches 59.03\%, with Conformer peaking at 65.10\% and SW reaching 73.44\% of subjects for the Conformer backbone. 
This cross-dataset, cross-architecture consistency---three distinct regularization strategies, three distinct backbone families, two datasets---confirms that respecting the psychological 
structure of emotion space is a general principle rather than an architecture- or method-specific artefact.


Moreover, subject-level analysis reveals widespread and consistent benefits across different individuals. As detailed in Table \ref{tab:seediv_scheme3}, our proposed regularization 
methods improve the F1 score for a majority of subjects. On the SEED-IV dataset, the SW method enhances performance for up to 76.67\% of subjects when applied to the AudioTransformer backbone. The improvements 
are even more pronounced on the challenging SEED-V dataset, where SW improves the F1 score for an average of 64.58\% of subjects across all three diverse backbones, peaking at 73.44\% for the Conformer model. 
These comprehensive results demonstrate that our framework not only boosts overall accuracy but also yields more robust, subject-generalized emotion 
representations that are less susceptible to individual differences in EEG signals.

The asymmetry in PV improvements between datasets is attributable to differences in graph topology. SEED-IV's graph is relatively dense for its size, limiting the number of misclassifications that qualify as non-adjacent violations. This compresses the dynamic range of PV.
In contrast, SEED-V's 5-class graph isolates Happy from the negative emotion cluster, connecting it only through Neutral, which 
creates more long-distance violation opportunities and makes PV a more sensitive quality indicator. Interpretability gains from graph regularization are therefore most salient 
when the emotion graph exhibits clear topological separation between clusters.

A complementary factor is annotation granularity. In both datasets, subjects provide a single emotion label for an entire trial (video clip), yet the model trains on short segments of a few 
seconds each, 
so segment-level targets inherit noise from trial-level aggregation. When labels imperfectly reflect the local emotion topology, 
the graph regularization signal is correspondingly weakened. SEED-V partially mitigates this through trial-level confidence scores, and the stronger improvements we observe on SEED-V may 
partly reflect this more reliable annotation protocol.

\section{Discussion and Conclusion}

We presented a graph-regularized framework for EEG emotion recognition that incorporates psychological emotion structure into the learning objective, penalizing cross-cluster confusions more heavily and constraining the hypothesis space to psychologically valid solutions that respects the intrinsci structure of emotion space. Three graph-based regularization strategies from different viewpoints via Graph Label Smoothing, Graph Laplacian with commuting distance, and Sliced Wasserstein distance were adapted and validated across three backbone architectures (Transformer, CNN-Transformer hybrid, GNN) on two benchmarks, our approach achieves up to $+5.42\%$ 
accuracy and $39\%$ reduction in clinically implausible errors, with consistent subject-level benefits in terms of upper bound of prediction capability. These results confirm that respecting emotion topology is an architecture-agnostic 
principle yielding predictions that are both more accurate and more interpretable. For practitioners, SW with adaptive weighting is recommended as the default: it achieved the best results on 4 of 6 model--dataset pairs and eliminates manual $\lambda$ tuning; GLS is preferred when compute is constrained. Limitations on the stopping criterion and the choice of $\lambda$ are acknowledged, and should be part of future work. 

\bibliographystyle{splncs04}
\bibliography{refs}
%




\end{document}